%% file: main.tex
\documentclass[10pt,twocolumn,letterpaper]{article}
\usepackage[utf8]{inputenc}

\usepackage{cvpr}
\usepackage{times}
\usepackage{epsfig}
\usepackage{graphicx}
\usepackage{amsmath}
\usepackage{amssymb}
\usepackage{mynotation}
\usepackage{algorithm}
\usepackage[noend]{algpseudocode}
\usepackage{graphicx}
\usepackage{booktabs}
\usepackage{pifont}
\usepackage{xcolor}
\usepackage[symbol]{footmisc}

\renewcommand{\thefootnote}{\fnsymbol{footnote}}

\algrenewcommand{\algorithmicrequire}{\textbf{Input:}}
\algrenewcommand{\algorithmicensure}{\textbf{Output:}}

\newcommand{\algcomment}[1]{\hfill{\footnotesize\# #1}}

\graphicspath{{./figures/}}

\usepackage[pagebackref=true,breaklinks=true,letterpaper=true,colorlinks,bookmarks=false]{hyperref}

\cvprfinalcopy 

\ifcvprfinal\fi
\begin{document}

\title{Learning Fast and Robust Target Models for Video Object Segmentation}

\newcommand{\sepa}{\hspace{20mm}}
\newcommand{\sepi}{\hspace{10mm}}
\author{Andreas Robinson$^1$\footnotemark[1] \sepa
	Felix Järemo Lawin$^1$\footnotemark[1]
	\sepa
	Martin Danelljan$^2$\vspace{1mm}\\
	Fahad Shahbaz Khan$^{1,3}$
	\sepa
	Michael Felsberg$^1$\vspace{2mm}\\
$^1$CVL, Link\"oping University, Sweden \sepi
$^2$CVL, ETH Zurich, Switzerland \sepi
$^3$IIAI, UAE
}
\maketitle

\footnotetext[1]{Authors contributed equally.}

\thispagestyle{empty}

\begin{abstract}
	Video object segmentation (VOS) is a highly challenging problem since the initial mask, defining the target object, is only given at test-time. The main difficulty is to effectively handle appearance changes and similar background objects, while maintaining accurate segmentation. Most previous approaches fine-tune segmentation networks on the first frame, resulting in impractical frame-rates and risk of overfitting. More recent methods integrate generative target appearance models, but either achieve limited robustness or require large amounts of training data.
	
	We propose a novel VOS architecture consisting of two network components. The target appearance model consists of a light-weight module, which is learned during the inference stage using fast optimization techniques to predict a coarse but robust target segmentation. The segmentation model is exclusively trained offline, designed to process the coarse scores into high quality segmentation masks. Our method is fast, easily trainable and remains highly effective in cases of limited training data. We perform extensive experiments on the challenging YouTube-VOS and DAVIS datasets. Our network achieves favorable performance, while operating at higher frame-rates compared to state-of-the-art. Code and trained models are available at \url{https://github.com/andr345/frtm-vos}.

\end{abstract}

\section{Introduction}

The problem of video object segmentation (VOS) has a variety of important applications, including object boundary estimation for grasping \cite{allen1993automated,kjellstrom2008visual}, autonomous driving \cite{ros2015vision,saleh2016kangaroo}, surveillance  \cite{cohen1999detecting,erdelyi2014adaptive} and video editing \cite{oh2018rgmp}. The task is to predict pixel-accurate masks of the region occupied by a specific target object, in every frame of a given video sequence. This work focuses on the semi-supervised setting, where a target ground truth mask is provided in the first frame. Challenges arise in dynamic environments with similar background objects and when the target undergoes considerable appearance changes or occlusions. Successful video object segmentation therefore requires both robust and accurate target pixel classification. 
\begin{figure}[t!]
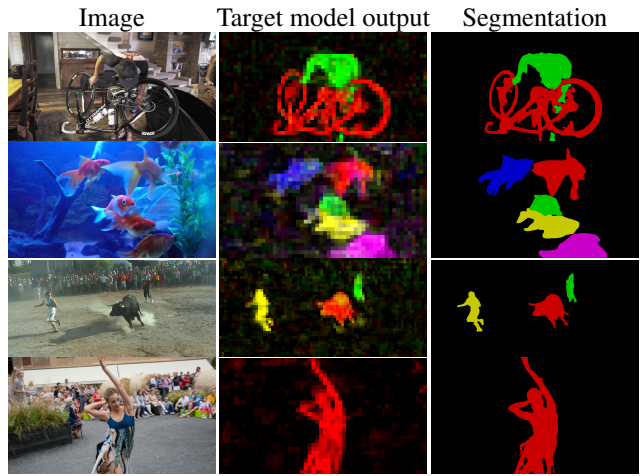

\newcommand{\image}{\includegraphics[width=0.33\columnwidth]}
\centering%
\tabcolsep=0.01cm%
\renewcommand{\arraystretch}{0.06}%
\begin{tabular}{ccc}
	{Image}  & {Target model output} & {\hspace{-3mm}Segmentation\hspace{-3mm}} \\
	\image{qresults/bike-packing/00034.jpg} &
	\image{qresults/bike-packing/00034-scores.png} &
	\image{qresults/bike-packing/00034.png} \\
	\image{qresults/gold-fish/00050.jpg} &
	\image{qresults/gold-fish/00050-scores.png} &
	\image{qresults/gold-fish/00050.png} \\
	\image{qresults/3b6c7988f6/00042.jpg} &
	\image{qresults/3b6c7988f6/00042-scores.png} &
	\image{qresults/3b6c7988f6/00042.png} \\
	\image{qresults/dance-twirl/00088.jpg} &
	\image{qresults/dance-twirl/00088-scores.png} &
	\image{qresults/dance-twirl/00088.png} \\
\end{tabular}
\vspace{1mm}
\caption{During inference, our target model learns to produce segmentation score maps (\textit{center}) of the target. As demonstrated in these examples, the target scores remain robust, despite difficult challenges, including distracting objects and appearance changes. Our segmentation network is trained to refine these coarse target score maps into a high-quality final object segmentation (\textit{right}).} 
\label{fig:intro}
\vspace{-4mm}
\end{figure}

Aiming to achieve a robust target-specific segmentation, several methods \cite{caelles2017osvos,maninis2017osvos_s,perazzi2017masktrack,xu2018youtube} fine-tune a generic segmentation network on the first frame, given the ground-truth mask. Although capable of generating accurate segmentation masks under favorable circumstances, these methods suffer from low frame-rates, impractical for many real world applications. Moreover, fine-tuning is prone to overfit to a single view of the scene, while degrading generic segmentation functionality learned during offline training. 
This limits performance in more challenging videos involving drastic appearance changes, occlusions and distractor objects~\cite{xu2018youtube2}. Further, the crucial fine-tuning step is not included in the offline training stage, which therefore does not simulate the full inference procedure.

Recent works~\cite{hu2018videomatch,johnander2018generative,oh2018rgmp,oh2019video,voigtlaender2018feelvos} address these limitations by employing internal models of the target and background appearance. They are based on, \eg, feature concatenation~\cite{oh2018rgmp}, feature matching~\cite{hu2018videomatch,oh2019video,voigtlaender2018feelvos} or Gaussian models~\cite{johnander2018generative}.
Such generative models have the advantage of facilitating efficient closed-form solutions that are easily integrated into neural networks. A drawback to these methods is the demand for large amounts of data in order to learn representations applicable for
the internal models~\cite{oh2018rgmp,oh2019video,wang2019ranet}. Due to the limited availability of annotated video data, these methods rely heavily on pre-training on image segmentation and synthesized VOS data via augmentation techniques. 
On the other hand, discriminative methods generally yield superior predictive power~\cite{ng2002discriminative} and have thus been preferred in many vision tasks, including image recognition~\cite{ILSVRC15}, object detection~\cite{lin2014microsoft} and tracking~\cite{VOT2018}. In this work, we therefore tackle the problem of integrating a \emph{discriminative} model of the target appearance into a VOS architecture.

Our approach integrates a light-weight discriminative target model and a segmentation network, for modeling the target appearance and generating accurate segmentation masks. Operating on deep features, the proposed target model learns during inference to provide robust segmentation scores. The segmentation network is designed to process features with the segmentation scores as guidance. During offline training the network learns to accurately adhere to object edges and to suppress erroneous classification scores from the target model, see Figure~\ref{fig:intro}. To learn the network parameters, we propose a training strategy that simulates the inference stage. This is realized by optimizing the target model on reference frames in each batch, and back-propagating the segmentation errors on corresponding validation frames. Contrary to fine-tuning based methods, the target adaption process is thus fully simulated during the offline training stage. During inference we keep the segmentation network fixed, while the target-specific learning is entirely performed by the target appearance model. Consequently, the segmentation network is target agnostic, retaining generic object segmentation functionality.

Unlike previous state-of-the-art methods, our discriminative target model requires no pre-training for image and synthetic video segmentation data. Our final approach, consisting of a single network architecture, is trained on VOS data in a single phase. Further, the employment of Gauss-Newton based optimization enables real-time video segmentation. 
We perform experiments on the DAVIS~\cite{perazzi2016davis} and YouTube-VOS 2018~\cite{xu2018youtube2} datasets and demonstrate the impact of the components of our proposed approach in an ablative analysis. We further compare our approach to several state-of-the-art methods. Despite its simplicity, our approach achieves an overall score of 76.7 on DAVIS 2017 and 72.1 on YouTube-VOS, while operating at 22 frames per second (FPS). We also evaluate a faster version of our approach that achieves a speed of 41 FPS, with only a slight degradation in segmentation accuracy.

\section{Related work}

The task of video object segmentation has seen extensive study and rapid development in recent years, largely driven by the introduction and evolution of benchmarks such as DAVIS \cite{perazzi2016davis} and YouTube-VOS \cite{xu2018youtube2}.

\noindent\textbf{First-frame fine-tuning:} Most state-of-the-art approaches train a segmentation network offline, and then fine-tune it on the first frame \cite{caelles2017osvos,maninis2017osvos_s,perazzi2017masktrack,xu2018youtube} to learn the target-specific appearance. This philosophy was extended \cite{voigtlaender2017onavos} by additionally fine-tuning on subsequent video frames. Other approaches \cite{cheng2017segflow,hu2018mgcrn,luiten2018premvos} further integrate optical flow as an additional cue. While obtaining impressive results on the DAVIS 2016 dataset, the extensive fine-tuning leads to impractically long run-times. Furthermore, such extensive fine-tuning is prone to overfitting, a problem only partially addressed by heavy data augmentation \cite{khoreva2017lucid}.

\noindent\textbf{Non-causal methods:}
Another line of research approaches the VOS problem by allowing non-causal processing~\cite{bao2018cinm,ci2018video,jang2017ctn,li2018dyenet}. In this work, we focus on the \emph{causal} setting in order to accommodate real-time applications.

\noindent\textbf{Mask propagation:} Several recent methods \cite{johnander2018generative,oh2018rgmp,oh2019video,perazzi2017masktrack,wang2019ranet,yang2018osnm} employ a mask-propagation module to improve spatio-temporal consistency of the segmentation. In \cite{perazzi2017masktrack}, the model is learned offline to predict the target mask through refinement of the previous frame's segmentation output. To further avoid first-frame fine-tuning, some approaches~\cite{oh2018rgmp, yang2018osnm} concatenate the current frame features with the previous mask and a target representation generated in the first frame. 
Unlike these methods, we do not explicitly enforce spatio-temporal consistency through mask-propagation. Instead, we use previous segmentation masks as training data for the discriminative model.

\noindent\textbf{Feature matching:} Recent methods \cite{chen2018blazingly,hu2018videomatch,oh2018rgmp,oh2019video,voigtlaender2018feelvos,vondrick2018tracking,wang2019ranet} incorporate feature matching to locate the target object. Rather than fine-tuning the network on the first frame, these methods first construct appearance models from features corresponding to the initial target labels. Features from incoming frames are then classified using techniques inspired by classical clustering methods \cite{chen2018blazingly,johnander2018generative} or feature matching  \cite{hu2018videomatch,voigtlaender2018feelvos,wang2019ranet}. In \cite{oh2019video}, a dynamic memory is used to combine feature matching from multiple previous frames.

\noindent\textbf{Tracking:}  
Efficient online learning of discriminative target-specific appearance models has been explored in visual tracking \cite{hare2016struck,henriques2015high}. Recently, optimization-based trackers \cite{DiMP,danelljan2018atom,danelljan2017eco} have achieved impressive results on benchmarks. These methods train convolution filters using efficient optimization to discriminate between target and background. The close relation between the two problem domains is made explicit in \cite{cheng2018favos}, where object trackers are used as external components to locate the target. Gauss-Newton has previously been used in object segmentation \cite{tjaden2018region} for pose estimation of known object shapes. In contrast, we do not employ off-the-shelf trackers to predict the target or rely on target pose estimation. Instead we take inspiration from the optimization-based learning of a discriminative model, in order to capture the target object appearance.

\input{method.tex}

\input{experiments.tex}

\section{Conclusion}

We propose video object segmentation approach, integrating a light-weight but highly discriminative target appearance model and a segmentation network. We find that despite its simplicity, a linear discriminative model is capable of generating robust target predictions. The segmentation network converts the predictions into high-quality object segmentations. The target model is efficiently trained during inference. 
Our method operates at high frame-rates and achieves state-of-the-art performance on the YouTube-VOS dataset and competitive results on DAVIS 2017 despite trained on limited data.

\noindent\textbf{Acknowledments:} This work was supported by the \mbox{ELLIIT} Excellence Center at Link\"oping-Lund for Information Technology, Autonomous Systems and Software Program (WASP) and the SSF project Symbicloud. 

{\small
	\bibliographystyle{ieee_fullname}
	\bibliography{references}
}
\clearpage
\input{supplement}

\end{document}

%% file: method.tex
\newcommand{\bI}{{\bf I}}  
\newcommand{\bK}{\mathcal{M}}  
\newcommand{\bx}{{\bf x}}  
\newcommand{\bs}{{\bf s}}  
\newcommand{\by}{{\bf y}}  
\newcommand{\bw}{{\bf w}}  
\newcommand{\bv}{{\bf v}}  

\section{Method}

In this work, we tackle the problem of predicting accurate segmentation masks of a target object, defined in the first frame of the video. This is addressed by constructing two network modules, $D$ and $S$, specifically designed for target modeling and segmentation respectively. The target model $D(\bx;\bw)$ takes features $\bx$ as input and generates a coarse, but robust, segmentation output $\bs=D(\bx;\bw)$ of the target object. It is parametrized by the weights $\bw$, which are solely learned during inference using the first-frame ground-truth, in order to capture the appearance of the target object. 

The coarse segmentation scores $\bs$, generated by the target model $D$, is passed to the segmentation network $S(\bs,\bx;\theta)$, additionally taking backbone features $\bx=F(I)$. The parameters $\theta$ of the segmentation network is only trained during the \emph{offline} training stage to predict the final high-resolution segmentation of the target. The coarse segmentation $\bs$ thus serves as a robust guide, indicating the target location. Crucially, this allows the segmentation network to remain target agnostic, and learn generic segmentation functionality. Since $S$ is trained with coarse segmentation inputs $\bs$ generated by the target model, it learns to enhance its prediction and correct mistakes.

\begin{figure}[h]
	\includegraphics[width=\columnwidth]{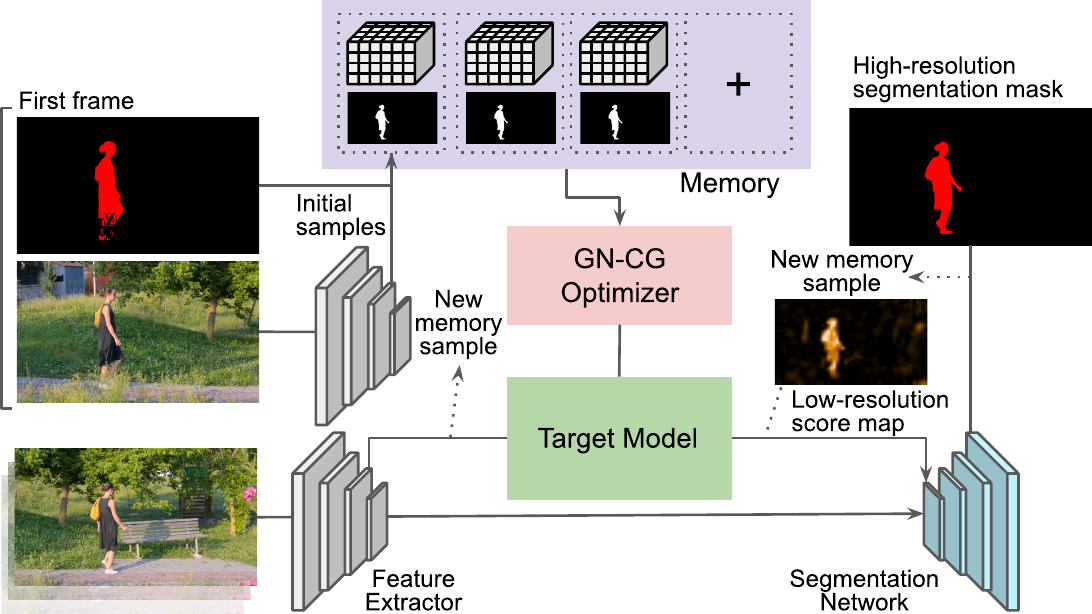}%
	\vspace{-1mm}%

	\caption{Overview of our video segmentation architecture. \textbf{Top left}: Feature maps are extracted from the first frame. \textbf{Top center}: The features and given ground-truth mask are then stored in memory, and used by the optimizer to train the target model. \textbf{Bottom left-center}: In subsequent frames, features are extracted and then classified as foreground or background by the target model, forming low-resolution score maps. \textbf{Bottom right}: The score maps are refined to high-resolution segmentation masks in the segmentation network. The high resolution masks and associated features are continuously added to the memory, and the target model is periodically updated by the optimizer. 
	}
	\label{fig:system}%
	\vspace{-3mm}
\end{figure}

During inference, we update the target model using the segmentation masks generated by $S$. Specifically, the mask and associated features are stored in a memory $\bK$. Before the next incoming frame, we further adapt our model to the target appearance by re-optimizing $D$ over all samples in $\bK$. In contrast to simply re-training on the latest frame, adding more training data to $\bK$ over time, reduces the risk for model drifting. Our full VOS architecture is illustrated in Figure~\ref{fig:system}.

\subsection{Target model}
\label{sec:discriminator}

We aim to develop a powerful and discriminative target appearance model, capable of differentiating between the target and background image regions. To successfully accommodate the VOS problem, the model must be robust to appearance changes and distractor objects. Moreover, it needs to be easily updated with new data and efficiently trainable. 
To this end, we employ a light-weight linear model $D(\bx; \bw)$ realized as two convolutional layers, 
\begin{equation}
\label{eq:target-model}
\bs = D({\bf x};{\bf w}) = {\bf w}_2 * ({\bf w}_1 * {\bf x}) \,,
\end{equation} 
with parameters $\bw$. These are trained exclusively during inference with image features $\bx$ and the target segmentation mask $\by$ given in the first video frame. It then takes input feature maps $\bx$ from subsequent video frames and outputs coarse segmentation scores $\bs$. The factorized formulation \eqref{eq:target-model} is used for efficiency, where the first layer $\bw_1$ reduces the feature dimensionality and the second layer $\bw_2$ computes the actual segmentation scores.

Fundamental to our approach, the target model parameters $\bw$ must be learned with minimal computational impact. To enable the deployment of fast converging optimization techniques, we adopt an $L^2$ loss given by,
\begin{equation}
\mathcal{L}_{D}({\bf w};\! \bK)\!=\!
\sum_k\!\gamma_k
\Vert \bv_k  \cdot  (\by_k - U(D(\bx_k)) \Vert^2 + \sum_j \!\lambda_j \Vert {\bf w}_j \Vert^2\!.
\label{eq:L2}
\end{equation}
Here, the parameters $\lambda_j$ control the regularization term and $\bv_k$ are weight masks balancing the impact of target and background pixels. $U$ denotes bilinear up-sampling of the output from the target model to the spatial resolution of the labels $\by_k$. The memory $\bK\! = \!\{(\bx_k, \by_k, \gamma_k)\}_{k=1}^{K}$ consists of sample feature maps $\bx_k$, target labels $\by_k$ and sample weights $\gamma_k$. During inference, $\bK$ is updated with new samples from the video sequence. To add more variety in the initial frame, we generate additional augmented samples $\{(\tilde{\bx}_k, \tilde{\by}_k, \gamma_k)\}_{k}$ using the initial image $I_0$ and labels $\by_0$, see supplement for more details. Compared to blindly updating on the latest frame \cite{voigtlaender2017onavos}, $\bK$ provides a controlled means of adding new samples while keeping past frames in memory by setting appropriate sample weights $\gamma_k$.

\noindent\textbf{Optimization:}
We employ the Gauss-Newton (GN) based strategy from \cite{danelljan2018atom} to optimize the parameters $\bw$. In comparison to the commonly used gradient descent based approaches, this strategy has significantly faster convergence properties \cite{marquardt1963algorithm}. In each iteration, the optimal increment $\Delta \bw$ is found using a quadratic approximation of the loss in \eqref{eq:L2}
\begin{equation}
\mathcal{L}_{D}(\bw +\Delta \bw)\! \approx \!\Delta \!\bw^T J_{\bw}^T J_{\bw}\Delta\bw + 2 \Delta \bw^T J_{\bw}^T r_{\bw} +r_{\bw}^T r_{\bw}\,\!.
\label{eq:gn}
\end{equation}
Here, $r_{\bw}$ contains the residuals \eqref{eq:L2} as $\sqrt{\gamma_k} \bv_k \cdot (\by_k - U(D(\bx_k)))$ and $\sqrt{\lambda_j} \bw_j$ and $J_{\bw}$ is the Jacobian of the residuals $r_{\bw}$ at $\bw$ and. The objective \eqref{eq:gn} results in a positive definite quadratic problem, which we minimize over $\Delta \bw$ with Conjugate Gradient (CG) descent \cite{hestenes1952methods}. We then update $\bw \leftarrow \bw+ \Delta \bw $ and execute the next GN iteration. 

\noindent\textbf{Pixel weighting:}
To address the imbalance between target and background, we employ a weight mask $\bv$ in \eqref{eq:L2} to ensure that the target influence is not too small relative to the usually much larger background region. We define the target influence as the fraction of target pixels in the image $\hat{\kappa}_k = N^{-1}\sum_{n} \by_k(n)$, where $n$ is the pixel index and $N$ the total number of pixels. The weight mask is then defined as
\begin{equation}
{\bf v}_k = 
\begin{cases}
\kappa/\hat{\kappa}_k, & ({\bf y}_k)_{n} = 1 \\
(1-\kappa) / (1-\hat{\kappa}_k), & ({\bf y}_k)_{n} = 0
\end{cases}
\end{equation}
where $\kappa = \max(\kappa_\mathrm{min}, \hat{\kappa}_k)$ is the desired and $\hat{\kappa}_k$ the actual target influence. We set $\kappa_\mathrm{min} = 0.1$ in our approach.

\subsection{Segmentation Network}
\label{sec:refinement}
\begin{figure}[t]
	\includegraphics[width=0.99\columnwidth]{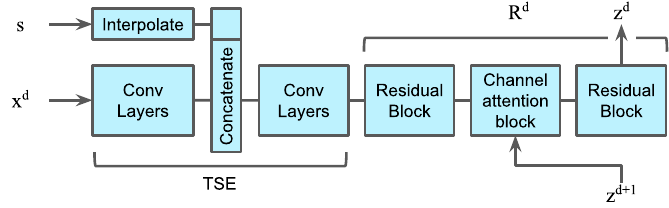}%
	\vspace{-2mm}%
	\caption{Illustration of one block in our segmentation network $S$. Target scores $\bs$ are resized to matching stride and merged with the backbone feature $\bx^d$, in the target segmentation encoder (TSE). The output from the TSE is further processed by a residual block before it is combined with the output from the deeper segmentation block in the CAB module. Finally, the output from the CAB module is processed by another residual block and sent to the next segmentation block. The complete network has four such layers.}
	\label{fig:network_block}
	\vspace{-3mm}
\end{figure}

While the target model provides robust but coarse segmentation scores, the final aim is to generate an accurate segmentation mask of the target at the original image resolution. To this end, we introduce a segmentation network, that processes the coarse score $\bs$ along with backbone features. The network consists of two types of building blocks: a target segmentation encoder (TSE) and a refinement module (see Figure~\ref{fig:network_block}).
From these we construct a U-Net based architecture for object segmentation as in \cite{yu2018dfn}. Unlike most state-of-the-art methods for semantic segmentation \cite{chen2018deeplab,zhao2017pyramid}, the U-Net structure does not rely on dilated convolutions, but effectively integrates low-resolution deep feature maps. This is crucial for reducing the computational complexity of our target model during inference. 

The segmentation network takes features maps $\bx^d$ as input from multiple depths in the backbone feature extractor network, with decreased resolution at each depth $d$. For each layer, $\bx^d$ along with the coarse scores $\bs$ are first processed by a TSE block $T^d$. The refinement module $R^d$ then inputs the resulting segmentation encoding generated by $T^d$ and the refined outputs ${\bf z}^{d+1}$ from the preceding deeper layer ${\bf z}^d = R^d(T^d(\bx^{d}, \bs) , {\bf z}^{d+1})$.
The refinement modules are comprised of two residual blocks and a channel attention block (CAB), as in \cite{yu2018dfn}. For the deepest block we set ${\bf z}^{d+1}$ to an intermediate projection of $\bx^d$ inside $T^d$. The output ${\bf z}^1$ at the shallowest layer is processed by two convolutional layers, providing the final refined segmentation output $\hat{\by}$. 

\noindent\textbf{Target segmentation encoder:} Seeking to integrate features and scores, we introduce the target segmentation encoder (TSE). It processes features in two steps, as visualized in Figure~\ref{fig:network_block} (right). First, we project the backbone features to 64 channels to reduce the subsequent computational complexity. We maintain 64 channels throughout the segmentation network, keeping the number of parameters low. After projection, the features are concatenated with the segmentation score $\bs$ and encoded by three convolutional layers.
\subsection{Offline Training}
\label{sec:offline}
We learn the parameters in our segmentation network offline by training on VOS training data. To this end, we propose a training scheme to simulate the inference stage. The network is trained on samples consisting of one reference frame and one or more validation frames. These are all randomly selected from the same video sequence.
A training iteration is then performed as follows: We first optimize the target model weights $\bw$, described in Section \ref{sec:discriminator}, based on the reference frame. We then apply our full network, along with the learned target model, on the validation frames to predict the target segmentations. The parameters in the network are learned by back-propagating through the binary cross-entropy loss with respect to the ground-truth masks.

During offline training, we only learn the parameters of the segmentation network, and freeze the weights of the feature extractor.
Since the target model only receives backbone features, we can pre-learn and store the target model weights for each sequence. The offline training time is therefore not significantly affected by the learning of $D$. 

The network is trained in a single phase on VOS data. We select one reference frame and two validation frames per sample and train the segmentation network with the ADAM optimizer \cite{kingma2014adam}. We start with the learning rate $\alpha=10^{-3}$, moment decay rates $\beta_1=0.9, \beta_2=0.999$ and weight decay $10^{-5}$, and train for about $10^6$ iterations, split into 120 epochs. The learning rate is then reduced to $\alpha=10^{-4}$, and we train for another 60 epochs. With pre-learned target model weights, the training is completed in less than a day. 

\subsection{Inference}
\label{sec:inference}
\begin{algorithm}[t]
	\caption{Inference}
	\begin{algorithmic}[1]
		\Require Images $\bI_i$, target $\by_0$
		\State$\bK_0(I_0, \by_0) = \{(\tilde{\bx}_k, \tilde{\by}_k, \gamma_k)\}_{k=1}^{K}$ \algcomment{Init dataset., sec \ref{sec:discriminator}}
		\State ${\bf w_0} = \mathrm{optimize}( \mathcal{L}_D(\bw;\bK_0)$) \algcomment{Init $D$, sec \ref{sec:discriminator}}
		\For{ $i = 1, 2, \ldots$ }
		\State	$\bx_i = F(\bI_i)$ \algcomment{Extract features}
		\State	$\bs_i = D(\bx_i; \bw_{i-1})$ \algcomment{Predict target, sec \ref{sec:discriminator}}
		\State	${\bf \hat y}_i = S(\bx_i, \bs_i)$ \algcomment{Segment target, sec \ref{sec:refinement}}
		\State	$\bK_i = \mathrm{extend}(\bx_i, \hat{\by}_i\, \gamma_i;\bK_{i-1})$ \algcomment{Extend dataset}
		\If{$i\mod{t_s} = 0$} \algcomment{Update $D$ every $t_s$ frame}
		\State ${\bf w}_{i} = \mathrm{optimize}( \mathcal{L}_D(\bw;, \bK_i))$ 
		\EndIf
		\EndFor
	\end{algorithmic}
	\label{alg:deploy}
\end{algorithm}
During inference, we apply our video segmentation procedure as summarized in Algorithm~\ref{alg:deploy}. We first generate the augmented dataset $\bK_0$, as described in Section~\ref{sec:discriminator}, given the initial image $\bI_0$ and the corresponding target mask $\by_0$. 
We then optimize the target model $D$ on this dataset. In the consecutive frames, we first predict the coarse segmentation scores $\bs$ using $D$. Next, we refine $\bs$ with the network $S$ (Section \ref{sec:refinement}). The resulting segmentation output $\hat{\by}_i$ along with the input features $\bx_i$ are added to the dataset. Each new sample $(\bx_i, \hat{\by}_i, \gamma_i)$ is first given a weight $\gamma_i = (1-\eta)^{-1}\gamma_{i-1}$, with $\gamma_0 = \eta$. We then normalize the sample weights to sum to unity. The parameter $\eta<1$ controls the update rate, such that the most recent samples in the sequence are prioritized in the re-optimization of the target model $D$. For practical purposes, we limit the maximum capacity $K_{\mathrm{max}}$ of the dataset. When the maximum capacity is reached, we remove the sample with the smallest weight from $\bK_{i-1}$ before inserting a new one. In all our experiments we set $\eta = 0.1$ and $K_{\mathrm{max}}=80$.

During inference, we optimize the target model parameters $\bw_{1}$ and $\bw_{2}$ on the current dataset $\bK_{i}$ every $t_s$-th frame. For efficiency, we keep the first layer of the target model $\bw_{1}$ fixed during updates. Setting $t_s$ to a large value reduces the inference time and regularizes the update of the target model. On the other hand, it is important that the target model is updated frequently, for objects that undergo rapid appearance changes. In our approach we set $t_s=8$.
The framework supports multi object segmentation by employing a target model for each object and fuse the final refined predictions with softmax aggregation as in \cite{oh2018rgmp}. We only require one feature extraction per image, since the features $\bx_i$ are common for all target objects.

\subsection{Implementation details}
\label{sec:imp-details}

We implement our method in the PyTorch framework~\cite{paszke2017automatic} and use a ResNet~\cite{He2015}, pre-trained on ImageNet~\cite{ILSVRC15}, as the feature extractor $F$. Following the naming convention in Table~1 of~\cite{He2015}, we extract four feature maps from the outputs of the blocks \verb|conv2_x| through \verb|conv5_x|. The target model $D$ accepts features from \verb|conv4_x| and produces 1-channel score maps. Both the input features and output scores have a spatial resolution 1/16th of the input image.

\noindent\textbf{Target model:} The first layer $\bw_{1}$ has $1 \times 1$ kernels reducing input features to $c=96$ channels while $\bw_{2}$ has a $3 \times 3$ kernel with one output channel. During first-frame optimization, $\bw_{1}$ and $\bw_{2}$ are randomly initialized. Using the data augmentation (see the supplementary material), we generate a initial dataset $\bK_0$ of 5 image and label pairs. We then optimize $\bw_{1}$ and $\bw_{2}$ with the Gauss-Newton algorithm outlined in Section \ref{sec:discriminator} with $N_\text{GN}=5$ GN steps. We apply $N_\text{CG}=10$ CG iterations in all GN steps but the first one. Since the initialization is random, we reduce the number of iterations to $N_\text{CGi}=5$ in the first step. In the target model update step we use $N_\text{CGu}=10$ CG iterations, updating $\bw_{2}$ every $t_s=8$ frame, while keeping $\bw_{1}$ fixed. We employ the aforementioned settings with a ResNet-101 backbone in our final approach, denoted {\bf Ours} in the following sections. 

We additionally develop a fast version, named {\bf Ours-fast}, with a ResNet-18 backbone and fewer optimization steps. Specifically, we set $N_\text{GN}=4$, $N_\text{CGi}=5$, $N_\text{CG}=10$ and $N_\text{CGu}=5$.

%% file: experiments.tex
\section{Experiments}

\newcommand{\mcJ}{\mathcal{J}}  
\newcommand{\mcF}{\mathcal{F}}  
\newcommand{\mcG}{\mathcal{G}} 
\newcommand{\mcJF}{\mathcal{J \& F}} 
We perform experiments on three benchmarks: DAVIS 2016 \cite{perazzi2016davis}, DAVIS 2017 \cite{perazzi2016davis} and YouTube-VOS \cite{xu2018youtube2}. For YouTube-VOS, we compare on the official validation set, with withheld ground-truth. For ablative experiments, we also show results on a separate validation split of the YouTube-VOS train set, consisting of 300 videos not used for training. Following the standard DAVIS protocol, we report both the mean Jaccard $\mcJ$ index and mean boundary $\mcF$ scores, along with the overall score $\mcJF$, which is the mean of the two. For comparisons on YouTube-VOS, we report $\mcJ$ and $\mcF$ scores for classes included in the training set (seen) and the ones that are not (unseen). The overall score $\mcG$ is computed as the average over all four scores, defined in YouTube-VOS. 
In addition, we compare the computational speed of the methods in terms of frames per second (FPS), computed by taking the average over the DAVIS 2016 validation set. For our approach, we used a V100 GPU and included all steps in Algorithm~\ref{alg:deploy} to compute the frame rates. Further results and analysis are provided in the supplement. 

\subsection{Ablation study}  

We analyze the contribution of the key components in our approach. All compared approaches are trained using the YouTube-VOS training split.

\noindent\textbf{Base net:} We construct a baseline network to analyze the impact of our target model $D$. This is performed by replacing $D$ with an offline-trained target encoder, and retraining the segmentation network $S$. As for our proposed network we keep the backbone $F$ parameters fixed. The target encoder is comprised of two convolutional layers, taking reference frame features from ResNet blocks \verb|conv4_x| and the corresponding target mask as input. Features (\verb|conv4_x|) extracted from the test frame are concatenated with the output from the target encoder and processed with two additional convolutional layers. The output is then passed to the segmentation network $S$ in the same manner as for the coarse segmentation score $\bs$ (see Section~\ref{sec:refinement}). We train this model with the same methodology as for our network. 

\noindent\textbf{F.-T:} We integrate a first-frame fine-tuning strategy into our network to compare this to our discriminative target model. For this purpose, we create an initial dataset $\bK_0$ with 20 samples using the \emph{same} sample generation procedure employed for our approach (section~\ref{sec:discriminator}). We then fine-tune all components of the network, except for the feature extractor, with supervision on the target model (loss in~\eqref{eq:L2}) and the pre-trained segmentation network (binary cross-entropy loss) using the ADAM optimizer with 100 iterations and a batch size of four. In this setting we omitted the proposed optimization strategy of the target model and instead initialize the parameters randomly before fine-tuning.

\noindent\textbf{$D$-only - no update:} To analyze the impact of the segmentation network $S$, we remove it from our architecture and instead let the target-specific model $D$ output the final segmentations. The coarse target model predictions are upsampled to full image resolution through bilinear interpolation. In this version, we only train the target model $D$ on the first frame, and refrain from subsequent updates.

\noindent\textbf{$D$-only:} We further enable target model updates (as described in Section \ref{sec:inference}) using the raw target predictions.

\noindent\textbf{Ours - no update:} For a fair comparison, we evaluate a variant of our approach with the segmentation network, but without any update of the target model $D$ during inference.

\noindent\textbf{Ours:} Finally, we include target model updates with segmentation network predictions to obtain our final approach.

In Table~\ref{tab:ab}, we present the results in terms of the $\mcJ$ score on a separate validation split of the YouTube-VOS training dataset. The base network, not employing the target model $D$, achieves a score of $49.8\%$. Employing fine-tuning on the first frame leads to an absolute improvement of $6\%$. Remarkably, using only the linear target model $D$ is on par with online fine-tuning. While fine-tuning an entire segmentation network is prone to severe overfitting to the initial frame, our shallow target model has limited capacity, acting as an implicit regularization mechanism that benefits robustness and generalization to unseen aspects of the target and background appearance.
Including updates results in an absolute improvement of $1.6\%$, demonstrates that we benefit from online updates despite the coarseness of the target mode generated labels. Further adding the segmentation network $S$ (Ours - no update) leads to a major absolute gain of $8.3\%$. This improvement stems from the offline-learned processing of the coarse segmentations, yielding more accurate mask predictions. Finally, the proposed online updating strategy additionally improves the score to $71.4\%$. 

\newcommand{\yes}{\checkmark}
\newcommand{\no}{\text{\ding{55}}}
\begin{table}[t!]
	\centering
	\resizebox{0.68\columnwidth}{!}{%
	\begin{tabular}[]{lcccc}
		\toprule
		Version & $D$ & $S$ & Update. & $\mcJ$ \\
		\midrule
		Base net & & \yes & &  49.8 \\
		F.-T. &  & \yes &  & 58.9 \\
		$D$-only -no update & \yes & & & 58.3  \\
		$D$-only & \yes & & \yes & 59.6  \\
		\textbf{Ours} - no update & \yes & \yes &   & 67.9 \\
		\textbf{Ours} & \yes & \yes & \yes &  \textbf{71.4}\\
		\bottomrule
	\end{tabular}
}\vspace{1mm}
	\caption{Ablative study on a validation split of 300 sequences from the YouTube-VOS train set. We analyze the different components of our approach, where $D$ and $S$ denote the target model and segmentation network respectively. Further, ``Update'' indicates if the target model update is enabled. Our target model $D$ outperforms the Base net and is comparable to first-frame fine-tuning (``F.-T.'') even with updates are disabled. Further, the segmentation network significantly improves the raw predictions from the target model $D$. Finally, the best performance is obtained when additionally updating target model $D$.}
	\label{tab:ab}
	\vspace{-4mm}
\end{table}

\subsection{Comparison to state-of-the-art}

We compare our method to recent approaches on the YouTube-VOS, DAVIS 2017 and DAVIS 2016 benchmarks. We provide results for two versions of our approach: \textbf{Ours} and \textbf{Ours (fast)} (see Section~\ref{sec:imp-details}). Many compared methods include additional training data or employ models that have been pre-trained on segmentation data. For fair comparison we classify methods into two categories: ``seg" for methods employing segmentation networks, pre-trained on e.g PASCAL~\cite{Everingham15} or MS-COCO~\cite{lin2014microsoft} and ``synth'' for methods that perform additional training on synthetic VOS data generated from image segmentation datasets.

\noindent\textbf{YouTube-VOS}~\cite{xu2018youtube2}: The official YouTube-VOS validation dataset has 474 sequences with objects from 91 classes. Out of these, 26 classes are not present in the training set. We provide results for {\bf Ours} and {\bf Ours (fast)}, both trained on the YouTube-VOS 2018 training set. We compare our method with the results reported in \cite{xu2018youtube}, that were obtained by retraining the methods on YouTube-VOS. Additionally, we compare to PReMVOS\cite{luiten2018premvos}, AGAME \cite{johnander2018generative}, RVOS~\cite{ventura2019rvos} and STM~\cite{oh2019video}. The results are reported in Table~\ref{tab:ytvos_results}. 

Among the methods using additional training data, OSVOS~\cite{caelles2017osvos}, OnAVOS~\cite{voigtlaender2017onavos} and PReMVOS employ first-frame fine-tuning, leading to inferior frame-rates below $0.3$ FPS. In addition to fine-tuning, PReMVOS constitutes a highly complex framework, encompassing multiple components and cues: mask-region proposals, optical flow based mask predictions, re-identification, merging and tracking modules. In contrast, our approach is simple, consisting of a single network together with a light-weight target model. Remarkably, our approach significantly outperforms PReMVOS by a relative margin of $5.2\%$, yielding a final $\mathcal{G}$-score of $72.1$. The recent STM method has the highest performance, employing feature matching with a dynamic memory to predict the target.

RVOS is trained only on YouTube-VOS, achieving a $\mathcal{G}$-score of $56.8$ by employing recurrent networks. In addition to recurrent networks, S2S employs first-frame fine-tuning, achieving a $\mathcal{G}$-score of $64.4$ with a significantly slower frame-rate compared to RVOS. In AGAME a generative appearance model is employed, resulting in a $\mathcal{G}$-score of $66.1$. We further report results from a version of STM (YV18), where training has been performed solely on YouTube-VOS. This significantly degrades the performance to a $\mathcal{G}$-score of $68.2$. {\bf Ours} outperforms all previous methods when only video data from YouTube-VOS has been used for training. We believe that, since our target model already provides robust predictions of the target on its own, our approach can achieve high performance without extensive training on additional data. Notably, {\bf Ours-fast}, maintains an impressive $\mathcal{G}$-score of $65.7$, while being significantly faster than all previous methods at $41.3$ FPS.

\begin{table}[t!]
\centering%
\resizebox{1.01\columnwidth}{!}{%
\input{yt2018_results_ordered_dset.tex}

}\vspace{1mm}
\caption{State-of-the-art comparison on the large-scale YouTube-VOS validation dataset, containing 474 videos. The results of our approach were obtained through the official evaluation server. We report the mean Jaccard ($\mcJ$) and boundary ($\mcF$) scores for object classes that are \emph{seen} and \emph{unseen} in the training set, along with the overall mean ($\mathcal{G}$). ``seg'' and ``synth'' indicate whether pre-trained segmentation models or additional data has been used during training. Our approaches achieve superior performance to methods that only trains on the YouTube-VOS train split, while operating at high frame-rates. Furthermore, {\bf Ours-fast} obtains the highest frame-rates while performing comparable to state-of-the-art.}
\label{tab:ytvos_results}
\vspace{-4mm}
\end{table}

\noindent\textbf{DAVIS 2017}~\cite{perazzi2016davis}: The validation set for DAVIS 2017 contains 30 sequences. We provide results for {\bf Ours} and {\bf Ours-fast}, trained a combination of the YouTube-VOS and DAVIS 2017 train splits, such that DAVIS 2017 is traversed eight times per epoch, and YouTube-VOS once. We report the results on DAVIS 2017 in Table~\ref{tab:dvjoint_results}. As in the YouTube-VOS comparison above, we categorize the methods with respect to usage of training data. Since our approaches ({\bf Ours} and {\bf Ours-fast}) and AGAME~\cite{johnander2018generative}, RVOS~\cite{ventura2019rvos}, STM~\cite{oh2019video} and FEELVOS~\cite{voigtlaender2018feelvos} all include the YouTube-VOS during training, we add a third category denoted ``yv".

OnAVOS~\cite{voigtlaender2017onavos}, OSVOS-S~\cite{maninis2017osvos_s}, MGCRN~\cite{hu2018mgcrn}, PReMVOS~\cite{luiten2018premvos} employ extensive fine-tuning on the first-frame, experiencing impractical segmentation speed. The methods RGMP~\cite{oh2018rgmp}, AGAME~\cite{johnander2018generative}, RANet~\cite{wang2019ranet} and FEELVOS~\cite{voigtlaender2018feelvos} all employ mask-propagation, which is combined with feature matching in the latter three methods. {\bf Ours} outperforms these methods with an $\mathcal{J \& F}$ score of $76.7$. In addition, {\bf Ours-fast} is significantly faster than all previous approaches, maintaining a $\mathcal{J \& F}$ score of $70.2$. 
Our method is only outperformed by PReMVOS and the recent STM, achieving $\mathcal{J \& F}$ scores of $77.8$ and $81.8$ respectively. PReMVOS, however, suffer from extremely slow frame rates: approximately 500 times lower than ours. Moreover, our approach outperforms PReMVOS on the more challenging and large scale YouTube-VOS (Table~\ref{tab:ytvos_results}) by a large margin.

We also evaluate our approach when only trained on DAVIS 2017, denoted {\bf Ours} (DV17). We compare this approach to the methods FAVOS~\cite{cheng2018favos}, AGAME (DV17)~\cite{johnander2018generative} and STM  (DV17)~\cite{oh2019video}, which have also only been trained on the DAVIS 2017 train split. Our method significantly outperform all these methods with a $\mathcal{J \& F}$ score of $68.8$. Moreover, this result is superior to OnAVOS, OSVOS-S, RANet, RVOS, RGMP and comparable to AGAME and FEELVOS despite their use of additional training data.

\begin{table}[t]
	\centering
	\tabcolsep=0.1cm
	\resizebox{0.9\columnwidth}{!}{%
	\input{dv_joint_results2.tex}
	}\vspace{1mm}
	\caption{State-of-the-art comparison on DAVIS 2017 and DAVIS 2016 validation sets. The columns with ``yv", ``seg", and ``synth" indicate whether YouTube-VOS, pre-trained segmentation models or additional synthetic data has been used during training. The best and second best entries are shown in red and blue respectively. In addition to {\bf Ours} and {\bf Ours-fast}, we report the results of our approach when trained on \emph{only} DAVIS 2017, in {\bf Ours} (DV17). Our approach outperform compared methods with practical frame-rates. Furthermore, we achieve competitive results when trained with only DAVIS 2017, owing to our discriminative target model.}
	\vspace{-4mm}
	\label{tab:dvjoint_results}
\end{table}

\noindent\textbf{DAVIS 2016}~\cite{perazzi2016davis}: Finally, we evaluate our method on the 20 validation sequences in DAVIS 2016, corresponding to a subset of DAVIS 2017 and report the results in Table~\ref{tab:dvjoint_results}. Our methods perform comparable to the fine-tuning based approaches PReMVOS, MGCRN~\cite{hu2018mgcrn}, OnAVOS and OSVOS-S. Further, {\bf Ours} outperforms AGAME, RGMP, OSNM, FAVOS and FEELVOS. 

\begin{figure*}[!t]
	\centering 
	\newcommand{\image}{\includegraphics[width=0.141\textwidth]}
	\tabcolsep=0.01cm
	\renewcommand{\arraystretch}{0.2}
	\input{q2.tex}\vspace{1mm}
	\caption{Examples of three sequences from DAVIS 2017, demonstrating how our method performs under significant appearance changes compared to ground truth, OSVOS-S \cite{maninis2017osvos_s}, AGAME \cite{johnander2018generative}, FEELVOS \cite{voigtlaender2018feelvos} and RGMP \cite{oh2018rgmp}.}
	\label{fig:qualitative_results2}
	\vspace{-3mm}
\end{figure*}
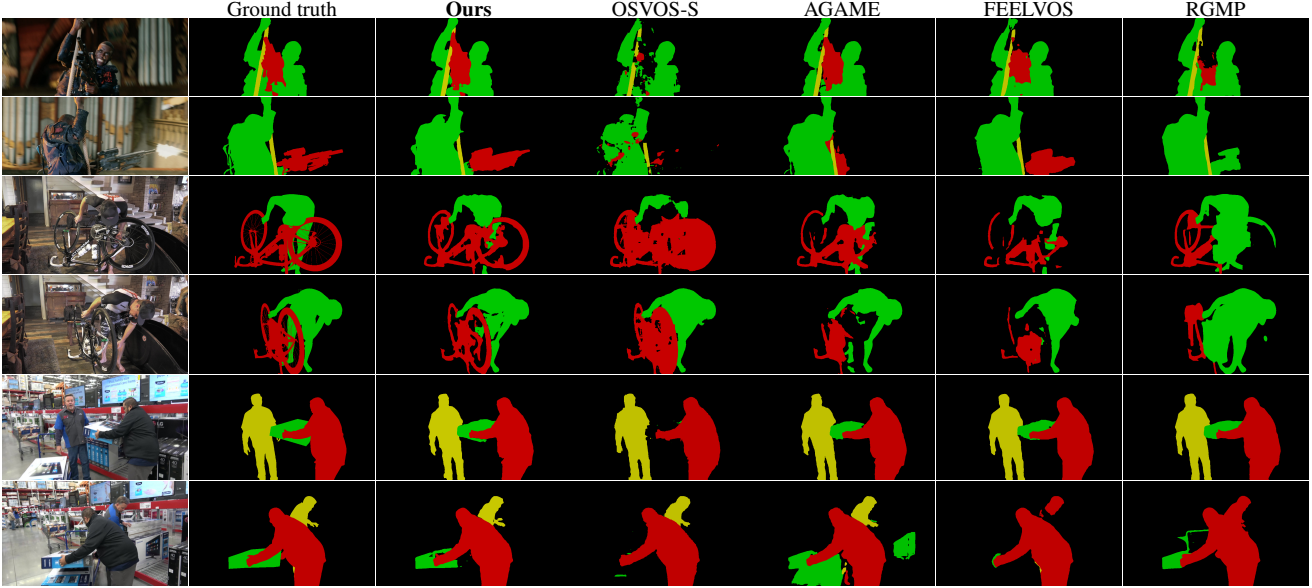

\begin{figure}
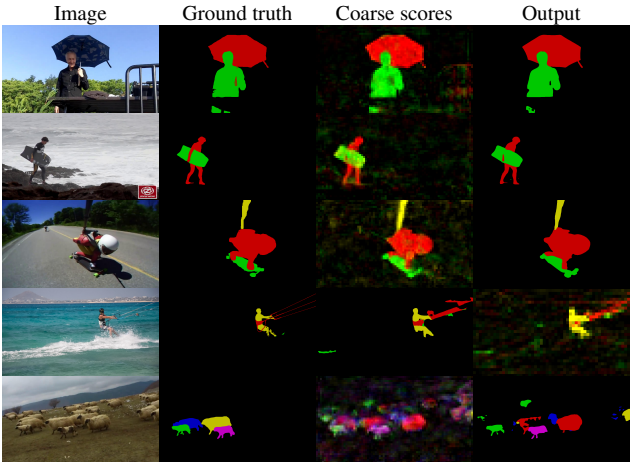

\newcommand{\image}{\includegraphics[width=0.248\columnwidth]}
\centering 
\tabcolsep=0.01cm
\renewcommand{\arraystretch}{0.06}
\begin{tabular}{cccc}

{\footnotesize Image} & {\footnotesize Ground truth} & {\footnotesize Coarse scores} & {\footnotesize Output}  \\

\image{qresults/14dae0dc93/00050.jpg} &
\image{qresults/14dae0dc93/00050-gt.png} &
\image{qresults/14dae0dc93/00050-scores.png} &
\image{qresults/14dae0dc93/00050.png} \\

\image{qresults/731b825695/00040.jpg} &
\image{qresults/731b825695/00040-gt.png} &
\image{qresults/731b825695/00040-scores.png}  &
\image{qresults/731b825695/00040.png} \\

\image{qresults/4f414dd6e7/00105.jpg} &
\image{qresults/4f414dd6e7/00105-gt.png} &
\image{qresults/4f414dd6e7/00105-scores.png} &
\image{qresults/4f414dd6e7/00105.png} \\

\image{qresults/kite-surf/00040.jpg} &
\image{qresults/kite-surf/00040-gt.png} &
\image{qresults/kite-surf/00040.png} &
\image{qresults/kite-surf/00040-scores.png} \\

\image{qresults/a9c9c1517e/00045.jpg} &
\image{qresults/a9c9c1517e/00045-gt.png} &
\image{qresults/a9c9c1517e/00045-scores.png} &
\image{qresults/a9c9c1517e/00045.png}  \\

\end{tabular}%
\vspace{0.5mm}%
\caption{Qualitative results of our method, showing both target model score maps and the output segmentation masks. The top three rows are success cases and the last two represents failures when objects are too thin or appear too similar. }
\label{fig:qualitative_results1}
\vspace{-2mm}
\end{figure}

\subsection{Qualitative Analysis}

\noindent\textbf{State-of-the-art:} We compare our approach to some state-of-the-art methods in Figure~\ref{fig:qualitative_results2}. By using an early and a late frame from each video sequence, we study how the methods cope with large target deformations over time.
In the first sequence (first and second row), the fine-tuning based OSVOS-S struggles as the target pose changes. While the mask-propagation in RGMP and generative appearance model in AGAME are accurate on fine details, they both fail to segment the red target, possibly due to occlusions. In the second and third sequences (rows three to six), all of the above methods fail to robustly segment the different targets. In contrast, our method accurately segments all targets in these challenging video sequences.

\noindent\textbf{Target model:} Some examples of the coarse segmentation scores and the final segmentation output are visualized in Figures~\ref{fig:intro} and ~\ref{fig:qualitative_results1}. In most cases, the target model provides robust segmentation scores of the target object. It however struggles is some cases where the target object contains thin or small structures or details. An example is the challenging kite lines in the \emph{kite-surfing} sequence, which are not accurately segmented.
This is likely due to the coarse feature maps the target model is operating on. It also have problems separating almost identical targets such as the sheep. On the other hand, the model successfully handles very similar targets as in the gold-fish sequence (row 2 in Figure~\ref{fig:intro}).

%% file: yt2018_results_ordered_dset.tex
\renewcommand{\yes}{\text{\ding{51}}}
\renewcommand{\no}{-}
\centering
\centering
\begin{tabular}{lcccccc}
	\toprule
	& \multicolumn{2}{c}{Training Data}& $\mathcal{G}$ & $\mathcal{J}$ & $\mathcal{F}$ & FPS\\
	Method & \footnotesize seg & \footnotesize synth & \footnotesize overall & \hspace{2.9mm}\footnotesize seen $|$ unseen & \footnotesize \hspace{2.2mm} seen $|$ unseen &  \footnotesize DAVIS16  \\
	\midrule
	{\bf Ours}      & \no &\no& {\bf72.1} & {\bf 72.3} $|$ {\bf 65.9} & {\bf 76.2} $|$ {\bf 74.1} & 21.9\\
	{\bf Ours-fast} & \no &\no& 65.7 & 68.6 $|$ 58.4 & 71.3 $|$ 64.5 & {\bf 41.3}\\
	STM (YV18) \cite{oh2019video}&\no&\no & 68.2 & - $|$ - & - $|$ - & 6.25\\
	AGAME \cite{johnander2018generative}&\no&\no & 66.1 & 67.8 $|$ 60.8 & 69.5 $|$ 66.2 & 14.3\\
	RVOS \cite{ventura2019rvos}&\no&\no & 56.8 & 63.6 $|$ 45.5 & 67.2 $|$ 51.0 & 22.7\\
	S2S \cite{xu2018youtube}&\no&\no & 64.4 & 71.0 $|$ 55.5 & 70.0 $|$ 61.2 & 0.11\\
	\midrule
	STM \cite{oh2019video}&\no&\yes& {\bf 79.4} & {\bf 79.7} $|$ {\bf 72.8} & {\bf 84.2} $|$ {\bf 80.9} & 6.25\\
	PReMVOS \cite{luiten2018premvos}&\yes & \yes & 66.9 & 71.4 $|$ 56.5 & - $|$ - & 0.03\\
	OnAVOS \cite{voigtlaender2017onavos}&\yes&\no & 55.2 & 60.1 $|$ 46.1 & 62.7 $|$ 51.4 & 0.08\\
	OSVOS \cite{caelles2017osvos}&\yes&\no & 58.8 & 59.8 $|$ 54.2 & 60.5 $|$ 60.7 & 0.22\\
	\bottomrule
\end{tabular}

%% file: dv_joint_results2.tex
\renewcommand{\yes}{\text{\ding{51}}}
\renewcommand{\no}{-}
\newcommand{\first}[1]{\textbf{\textcolor{red}{#1}}}
\newcommand{\second}[1]{\textit{\textcolor{blue}{#1}}}
\begin{tabular}{@{~}lccc@{~~~}c|c|c}
\toprule
       & \multicolumn{3}{c@{~~~}}{Training Data} & DAVIS17  & DAVIS16 &  \\
Method & yv & seg & synth & $\mcJF$ & $\mcJF$ & FPS \\
\midrule
{\bf Ours} (DV17)                          & \no & \no & \no & \first{68.8} & \first{81.7} & \first{21.9} \\
AGAME (DV17) \cite{johnander2018generative}& \no & \no & \no & \second{63.2} & -    & \second{14.3} \\
FAVOS \cite{cheng2018favos}                & \no & \no & \no & 58.2 & 80.8 & 0.56 \\
STM (DV17) \cite{oh2019video}              & \no & \no & \no & 43.0 & -    & 6.25 \\
\midrule
{\bf Ours}                              & \yes& \no & \no & 76.7 & 83.5 & 21.9 \\
{\bf Ours-fast}                         & \yes& \no & \no & 70.2 & 78.5 & \first{41.3} \\
RVOS \cite{ventura2019rvos}                & \yes& \no & \no & 50.3 & -    & 22.7 \\
RANet \cite{wang2019ranet}                 & \no & \no &\yes & 65.7 & 85.5 & \second{30.3} \\
AGAME \cite{johnander2018generative}       & \yes& \no &\yes & 70.0 & -    & 14.3 \\
STM \cite{oh2019video}                     & \yes& \no &\yes & \first{81.8} & \first{89.3} & 6.25 \\
RGMP \cite{oh2018rgmp}                     & \no & \no &\yes & 66.7 & 81.8 & 7.7  \\
FEELVOS \cite{voigtlaender2018feelvos}     & \yes&\yes & \no & 71.5 & 81.7 & 2.22 \\
OSNM \cite{yang2018osnm}                   & \no & \no &\yes & 54.8 & -    & 7.14 \\
PReMVOS \cite{luiten2018premvos}           & \no &\yes &\yes & \second{77.8} & \second{86.8}  & 0.03 \\
OSVOS-S \cite{maninis2017osvos_s}          & \no &\yes & \no & 68.0         & 86.5 & 0.22 \\
OnAVOS \cite{voigtlaender2017onavos}       & \no &\yes & \no & 67.9         & 85.5          & 0.08 \\
MGCRN \cite{hu2018mgcrn}                   & \no &\yes & \no & -            & 85.1          & 1.37 \\
\bottomrule
\end{tabular}

%% file: q2.tex
\begin{tabular}{ccccccc}
 & {\footnotesize Ground truth} & {\footnotesize {\bf Ours}} & {\footnotesize OSVOS-S} & {\footnotesize AGAME} & {\footnotesize FEELVOS} & {\footnotesize RGMP} \\
\image{qresults/shooting/00004.jpg} &
\image{qresults/shooting/00004-gt.png} &
\image{qresults/shooting/00004.png} &
\image{qresults/shooting/00004-osvoss.png} &
\image{qresults/shooting/00004-agame.png} &
\image{qresults/shooting/00004-feelvos.png} &
\image{qresults/shooting/00004-rgmp.png} \\
\image{qresults/shooting/00039.jpg} &
\image{qresults/shooting/00039-gt.png} &
\image{qresults/shooting/00039.png} &
\image{qresults/shooting/00039-osvoss.png} &
\image{qresults/shooting/00039-agame.png} &
\image{qresults/shooting/00039-feelvos.png} &
\image{qresults/shooting/00039-rgmp.png} \\

\image{qresults/bike-packing/00034.jpg} &
\image{qresults/bike-packing/00034-gt.png} &
\image{qresults/bike-packing/00034.png} &
\image{qresults/bike-packing/00034-osvos.png} &
\image{qresults/bike-packing/00034-agame.png} &
\image{qresults/bike-packing/00034-feelvos.png} &
\image{qresults/bike-packing/00034-rgmp.png} \\
\image{qresults/bike-packing/00068.jpg} &
\image{qresults/bike-packing/00068-gt.png} &
\image{qresults/bike-packing/00068.png} &
\image{qresults/bike-packing/00068-osvos.png} &
\image{qresults/bike-packing/00068-agame.png} &
\image{qresults/bike-packing/00068-feelvos.png} &
\image{qresults/bike-packing/00068-rgmp.png} \\

\image{qresults/loading/00005.jpg} &
\image{qresults/loading/00005-gt.png} &
\image{qresults/loading/00005.png} &
\image{qresults/loading/00005-osvoss.png} &
\image{qresults/loading/00005-agame.png} &
\image{qresults/loading/00005-feelvos.png} &
\image{qresults/loading/00005-rgmp.png} \\
\image{qresults/loading/00049.jpg} &
\image{qresults/loading/00049-gt.png} &
\image{qresults/loading/00049.png} &
\image{qresults/loading/00049-osvoss.png} &
\image{qresults/loading/00049-agame.png} &
\image{qresults/loading/00049-feelvos.png} &
\image{qresults/loading/00049-rgmp.png} \\
\end{tabular}

%% file: supplement.tex
\renewcommand{\bI}{{\bf I}}  
\renewcommand{\bK}{\mathcal{M}}  
\renewcommand{\bx}{{\bf x}}  
\renewcommand{\bs}{{\bf s}}  
\renewcommand{\by}{{\bf y}}  
\renewcommand{\bw}{{\bf w}}  
\renewcommand{\bv}{{\bf v}}  

\begin{center}
\textbf{\large Supplementary Material}
\end{center}
In this supplementary material we provide a description of the initial sample generation, referred in section 3.1 in the paper. We also provide more detailed quantitative results and ablative analysis of parameters. 

\section{Initial sample generation}
\begin{figure}[h]
	\centering 
	\includegraphics[width=\columnwidth]{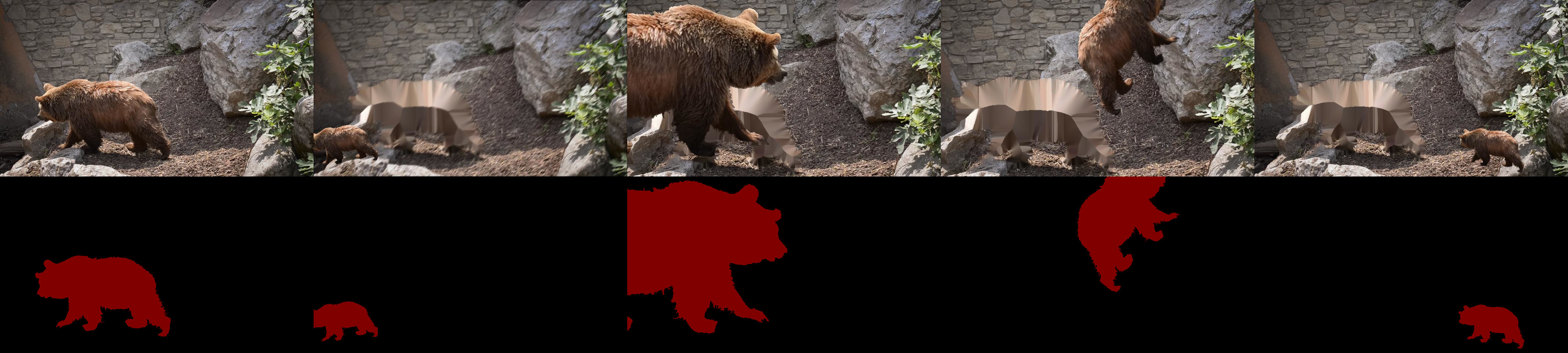}
	\caption{Example of data augmentation for the initial sample generation with the image (top row) and corresponding label mask (bottom row). The original first frame sample is shown to the left and the augmented samples follows from left to right.}
	\label{fig:data-aug-example}
\end{figure}

In the first frame of the sequence, we train the target model on the initial dataset $\bK_0$, created from the given target mask $\by_0$ and the extracted features $\bx_0$. To add more variety in the initial frame, we generate additional augmented samples. Based on the initial label $\by_0$, we cut out the target object and apply a fast inpainting method \cite{telea2004inpainting} to restore the background. We then apply a random affine warp and blur before pasting the target object back onto the image, creating a set of augmented images $\tilde{\bI}_k$ and corresponding label masks $\tilde{\by}_k$.  
After feature extraction, we insert the unmodified first frame and the augmented frames into the dataset $\bK_0 = \{(\tilde{\bx}_k, \tilde{\by}_k, \gamma_k)\}_{k=0}^{K-1}$ and set the sample weights $\gamma_k$ such that the original sample carries twice the weight of the other samples. Example augmentations performed in the initial frame are shown in Figure~\ref{fig:data-aug-example}.

\section{Detailed Quantitative Results}

In this section we report some additional quantitative results.

\subsection{Training data}
We analyze how the amount of training data impacts the performance of our approach. For this purpose we train our model on subsets of the YouTube-VOS training set containing 100\%, 50\%, 25\% and 0\% of the YouTube-VOS 2018 training split (excluding the validation split used to analyzing our approach as in Section~4 in the paper). For the version using 0\% of the data, called ``Ours D-only'', we only apply or target appearance model, which is trained during inference, thus requiring no offline training. As shown in Table~\ref{tab:t_data}, the performance improves as we increase the amount of training data from 0 to 100 percent of the YouTubeVOS training split.
Already at 25 percent our approach outperforms recent methods such as AGAME~\cite{johnander2018generative} (see Table~2 in the paper). At 50 percent, our approach surpasses all compared methods in Table~2 in the paper, that are trained only on the full YouTube-VOS training set. Remarkably, our target model  without the segmentation (Ours D-only), consisting of a linear filter that requires no pre-training, obtains a $\mathcal{G}$-score superior to the methods OSVOS~\cite{caelles2017osvos}, OnAVOS~\cite{voigtlaender2017onavos} and the recent RVOS~\cite{ventura2019rvos} (see Table~2 in the paper).

\begin{table}
\centering
\resizebox{0.95\columnwidth}{!}{%
\begin{tabular}{lcccc}
\toprule
       & $\mathcal{G}$ & $\mathcal{J}$ & $\mathcal{F}$ & \\
Method & \footnotesize overall & \footnotesize \hspace{2.0mm} seen $|$ unseen & \footnotesize \hspace{2.0mm} seen $|$ unseen &  \footnotesize data  \\
\midrule
Ours & 72.1 & 72.3 $|$ 65.9 & 76.2 $|$ 74.1 & 100\% \\
Ours & 70.6 & 71.4 $|$ 63.7 & 75.5 $|$ 71.8 & 50\% \\
Ours & 66.7 & 69.7 $|$ 58.5 & 73.0 $|$ 65.6 & 25\% \\
Ours D-only & 59.9 & 60.1 $|$ 57.0 & 58.6 $|$ 63.8 & 0\% \\

\bottomrule
\end{tabular}
}\vspace{1mm}
\caption{YouTubeVos 2018 test-dev results for different amount of training data, sample. Ours with 100\% data is the same instance as in the comparison in Table~2 in the main paper. The Ours D-only is our approach without the segmentation network as described in Section~5.1 in the main paper. It thus requires no training data at all.}
\label{tab:t_data}
\end{table}

\subsection{Algorithm runtime analysis}
We investigate the runtime for the different steps in our proposed VOS approach in Algorithm 1 in the main paper. All runtimes have been computed by averaging over the DAVIS 2016 evaluation split.

\begin{figure}[h]
	\centering 
	\includegraphics[width=0.49\textwidth]{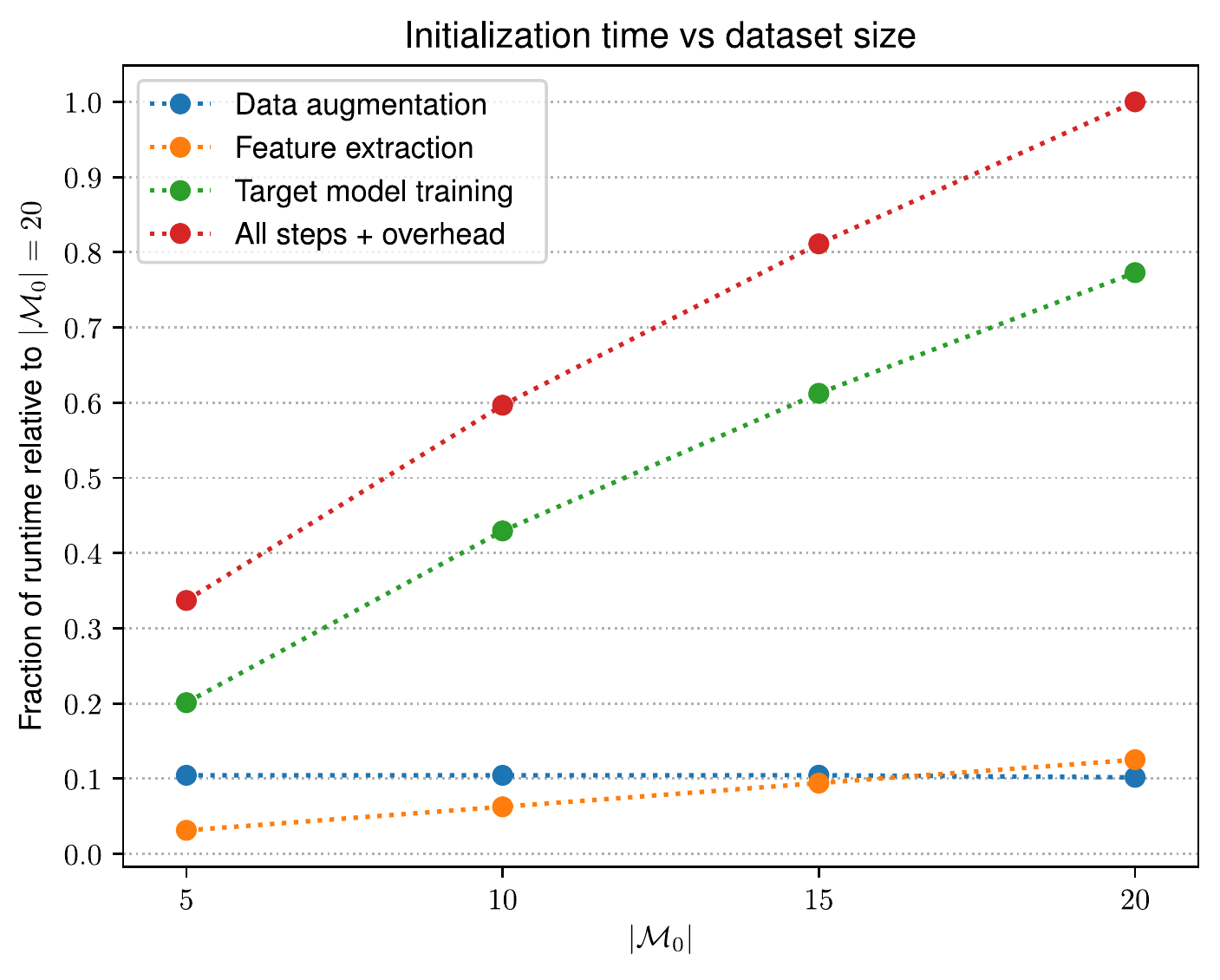}
	\caption{Initialization runtime (frame 1), relative to the initial dataset size $|\mathcal{M}_0| = 20$ with a fixed maximum dataset size  $K_\mathrm{max} = 80$ (section 3.4).}
	\label{fig:avg_init_runtime}
\end{figure}

Figure \ref{fig:avg_init_runtime} shows how execution in frame 1 (the \textit{init phase}, steps 1 and 2 in Algorithm 1) changes vs the size of the initial sample memory (or dataset) $\mathcal{M}_0$, when segmenting a single object. We present relative runtimes of the maximum time spent using all steps with $|\mathcal{M}_0|=20$ The time spent during data augmentation is dominated by the inpainting which is performed only once, on the first frame, and hence it is appears constant.

\begin{figure}[h]
	\centering 
	\includegraphics[width=0.49\textwidth]{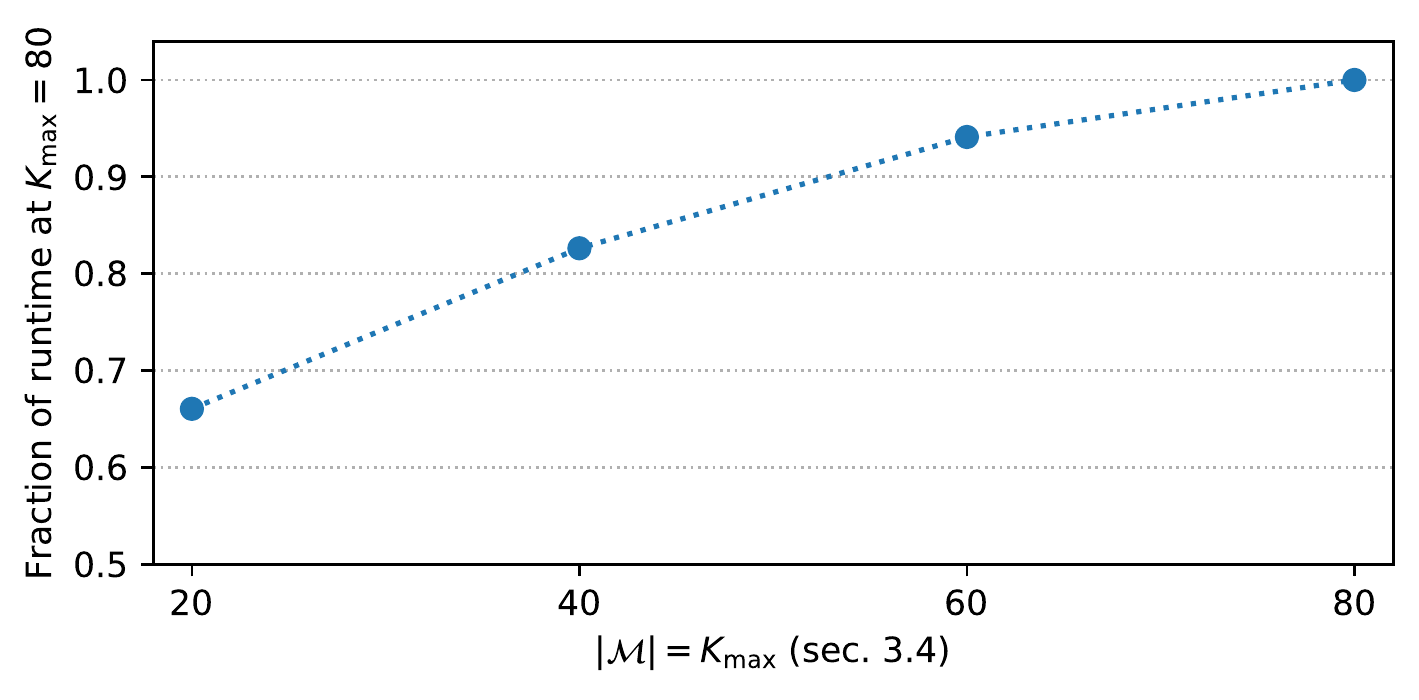}
	\caption{Per-frame (forward phase) runtime relative to the maximum dataset size $K_\mathrm{max} = 80$ with a fixed initial size $|\mathcal{M}_0| = 20$.}
	\label{fig:avg_fw_runtime}
\end{figure}

Figure \ref{fig:avg_fw_runtime} shows how execution in frames 2 and onward (the \textit{forward phase}, steps 4-9 in Algorithm 1) when fixating $|\mathcal{M}_0| = 20$ and varying the maximum dataset size $K_\mathrm{max}$. Again, we present timings in relation to the full execution time using $|\mathcal{M}|=80$. Note that most DAVIS 2016 videos are only a few seconds long and will never fill $\mathcal{M}$ entirely. This will reduce the apparent runtimes for large dataset sizes.

In addition, Table \ref{tab:fw_distr} shows the distribution of average time spent on each step in one frame in the forward phase. This is in the steady-state situation, after the sample memory is filled (here $|\mathcal{M}_i| = 80$), averaged over the last $t_s=8$ frames of the sequence.

Since the DAVIS videos are quite short, the init phase accounts for 41 percent of the total runtime when evaluating the {\bf Ours} variant on DAVIS2016. On a per video basis, the initialization requires between 31 (for ``cows'' with 104 frames) and 60 percent (for ``car-shadow'' with 40 frames) of the total runtime.

\begin{table}[h]
\centering
\begin{tabular}{lc}
    \toprule
    \textbf{Algorithm step} & \textbf{Percent} \\
     \midrule
     4. Feature extraction   & 24.0 \\
     6. Segmentation & 10.0 \\
     9. Target model update training & 65.5 \\
     Other & 0.6 \\
     \bottomrule
\end{tabular}
\caption{Distribution of time spent on steps in the frame loop of algorithm 1. The target prediction (step 5) is wrapped into ``Other". }
\label{tab:fw_distr}
\end{table}

From figure \ref{fig:avg_init_runtime}, we conclude that the first-frame initialization (algorithm steps 1-2) scales approximately linearly with $|\mathcal{M}_0|$. The per-frame (forward phase) processing (Algorithm 1 steps 3-9) is dominated by the model update training and feature extraction. Theoretically, the complexity of both phases scale linearly with the number of iterations in their respective optimization steps (step 2 and 9) as well as linearly with the number of targets.

\subsection{Parameter sensitivity}
\begin{figure}[h]
	\centering 
	\includegraphics[width=\columnwidth]{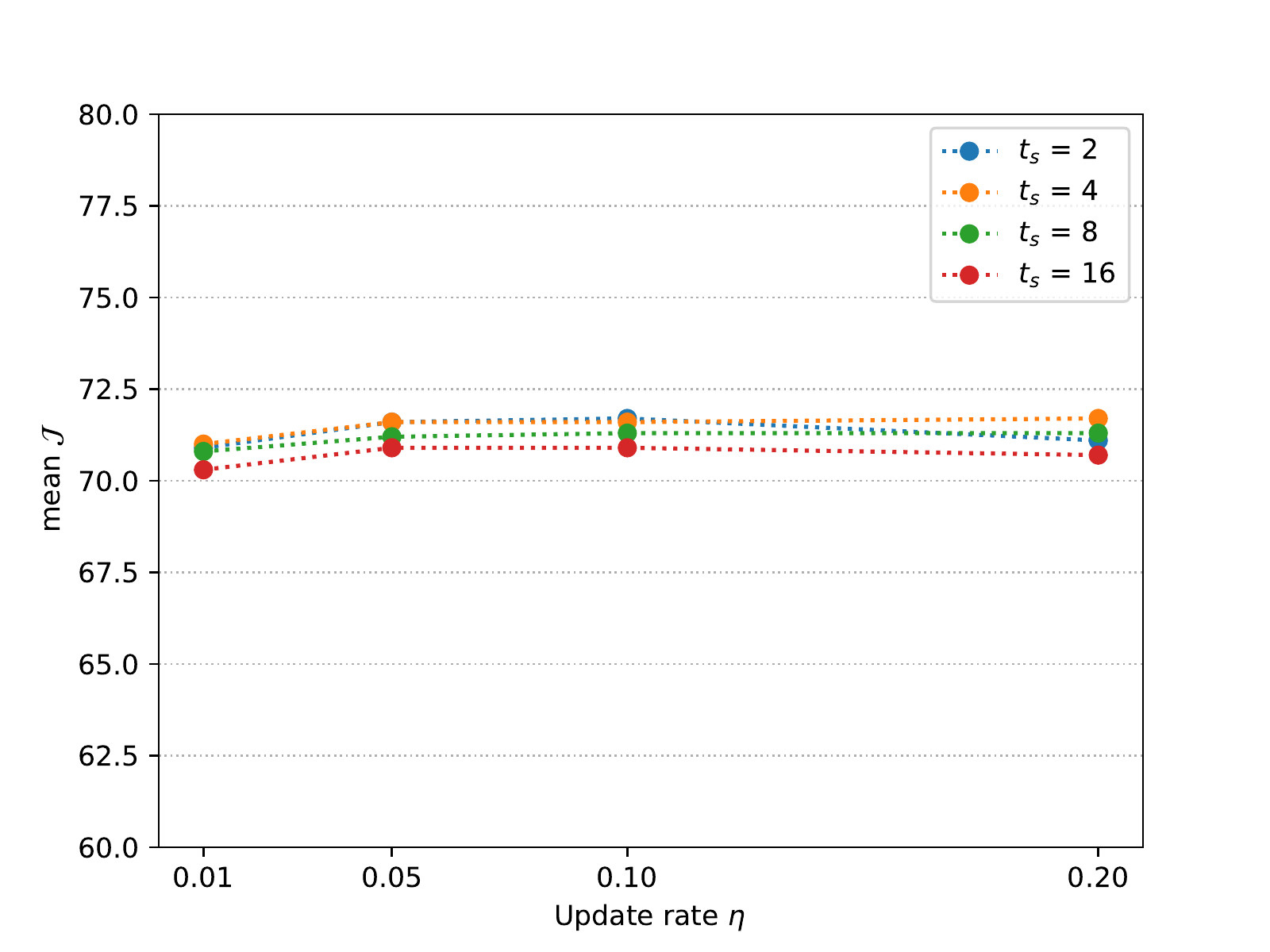}
	\caption{Mean intersection-over-union results on our YouTubeVOS validation split (defined in Section 4), as a function of the memory update rate $\eta$ and retraining interval $t_s$ hyper parameters, detailed in section 3.4. }
	\label{fig:lr_ts_grid}
\end{figure}

Figure \ref{fig:lr_ts_grid} reports the mean $\mathcal{J}$ as functions of the memory learning rate $\eta$ and target model update interval $t_s$  (defined in Section 3.4 in the paper). The experiments are performed on the YouTubeVOS validation split, defined in Section 4 in the paper. It is apparent that the method is rather insensitive to either parameter.

In addition, Table \ref{tab:num_aug_test} shows the mean $\mathcal{J}$ as functions of the size of the initial training dataset $\mathcal{M}_0$.
We test two variants of our method, one trained on YouTubeVOS data and one trained on both YouTubeVOS and DAVIS data. We evaluate on our own YouTubeVOS validation split and the DAVIS validation set. 
We observe that the YouTubeVOS evaluation is insensitive to the choice of $|\mathcal{M}_0|$. While still achieving a competitive $\mathcal{J}$-score without initial data augmentation, our approach obtains the best performance using four additional augmented samples in $\mathcal{M}_0$.

\begin{table}[h]
\centering
\begin{tabular}{cccccc}
    \toprule
                     &      & \multicolumn{4}{c}{$|\mathcal{M}_0|$} \\
    \textbf{Method}  & \textbf{Eval} & 1 & 5 & 10 & 20 \\
     \midrule
     Ours (yt)       & ytv & 71.5 & 71.4 & 71.2 & 71.4 \\
     Ours (yt+dv)    & dvv & 69.4 & 73.8 & 72.6 & 73.1 \\
     \bottomrule
\end{tabular}
\caption{The influence on mean $\mathcal{J}$ with varying $|\mathcal{M}_0|$ during inference. We test two variants, trained on either YouTubeVOS only (yt) or both YouTubeVOS and DAVIS2017 (yt+dv17). Results shown are from evaluating on our YoutubeVOS validation split (ytv) and the DAVS2017 validation split (dvv).}
\label{tab:num_aug_test}
\end{table}

%% file: main.bbl
\begin{thebibliography}{10}\itemsep=-1pt

\bibitem{allen1993automated}
Peter~K Allen, Aleksandar Timcenko, Billibon Yoshimi, and Paul Michelman.
\newblock Automated tracking and grasping of a moving object with a robotic
  hand-eye system.
\newblock {\em IEEE Transactions on Robotics and Automation}, 9(2):152--165,
  1993.

\bibitem{bao2018cinm}
Linchao Bao, Baoyuan Wu, and Wei Liu.
\newblock Cnn in mrf: Video object segmentation via inference in a cnn-based
  higher-order spatio-temporal mrf.
\newblock In {\em Proceedings of the IEEE Conference on Computer Vision and
  Pattern Recognition}, pages 5977--5986, 2018.

\bibitem{DiMP}
Goutam Bhat, Martin Danelljan, Luc~Van Gool, and Radu Timofte.
\newblock Learning discriminative model prediction for tracking.
\newblock In {\em {IEEE/CVF} International Conference on Computer Vision},
  pages 6181--6190, 2019.

\bibitem{caelles2017osvos}
Sergi Caelles, K-K Maninis, Jordi Pont-Tuset, Laura Leal-Taix{\'e}, Daniel
  Cremers, and Luc Van~Gool.
\newblock One-shot video object segmentation.
\newblock In {\em 2017 IEEE Conference on Computer Vision and Pattern
  Recognition (CVPR)}, pages 5320--5329. IEEE, 2017.

\bibitem{chen2018deeplab}
Liang-Chieh Chen, George Papandreou, Iasonas Kokkinos, Kevin Murphy, and Alan~L
  Yuille.
\newblock Deeplab: Semantic image segmentation with deep convolutional nets,
  atrous convolution, and fully connected crfs.
\newblock {\em IEEE transactions on pattern analysis and machine intelligence},
  40(4):834--848, 2018.

\bibitem{chen2018blazingly}
Yuhua Chen, Jordi Pont-Tuset, Alberto Montes, and Luc Van~Gool.
\newblock Blazingly fast video object segmentation with pixel-wise metric
  learning.
\newblock In {\em Proceedings of the IEEE Conference on Computer Vision and
  Pattern Recognition}, pages 1189--1198, 2018.

\bibitem{cheng2018favos}
J. Cheng, Y.-H. Tsai, W.-C. Hung, S. Wang, and M.-H. Yang.
\newblock Fast and accurate online video object segmentation via tracking
  parts.
\newblock In {\em IEEE Conference on Computer Vision and Pattern Recognition
  (CVPR)}, 2018.

\bibitem{cheng2017segflow}
Jingchun Cheng, Yi-Hsuan Tsai, Shengjin Wang, and Ming-Hsuan Yang.
\newblock Segflow: Joint learning for video object segmentation and optical
  flow.
\newblock In {\em 2017 IEEE International Conference on Computer Vision}, pages
  686--695. IEEE, 2017.

\bibitem{ci2018video}
Hai Ci, Chunyu Wang, and Yizhou Wang.
\newblock Video object segmentation by learning location-sensitive embeddings.
\newblock In {\em European Conference on Computer Vision}, pages 524--539.
  Springer, 2018.

\bibitem{cohen1999detecting}
I Cohen and G Medioni.
\newblock Detecting and tracking moving objects for video surveillance.
\newblock In {\em Proceedings. 1999 IEEE Computer Society Conference on
  Computer Vision and Pattern Recognition (Cat. No PR00149)}, volume~2, pages
  319--325. IEEE, 1999.

\bibitem{danelljan2018atom}
Martin Danelljan, Goutam Bhat, Fahad~Shahbaz Khan, and Michael Felsberg.
\newblock {ATOM}: Accurate tracking by overlap maximization.
\newblock In {\em IEEE Conference on Computer Vision and Pattern Recognition},
  2019.

\bibitem{danelljan2017eco}
Martin Danelljan, Goutam Bhat, Fahad Shahbaz~Khan, and Michael Felsberg.
\newblock {ECO}: Efficient convolution operators for tracking.
\newblock In {\em Proceedings of the IEEE Conference on Computer Vision and
  Pattern Recognition}, pages 6638--6646, 2017.

\bibitem{erdelyi2014adaptive}
Ad{\'a}m Erd{\'e}lyi, Tibor Bar{\'a}t, Patrick Valet, Thomas Winkler, and
  Bernhard Rinner.
\newblock Adaptive cartooning for privacy protection in camera networks.
\newblock In {\em 2014 11th IEEE International Conference on Advanced Video and
  Signal Based Surveillance (AVSS)}, pages 44--49. IEEE, 2014.

\bibitem{Everingham15}
M. Everingham, S.~M.~A. Eslami, L. Van~Gool, C.~K.~I. Williams, J. Winn, and A.
  Zisserman.
\newblock The pascal visual object classes challenge: A retrospective.
\newblock {\em International Journal of Computer Vision}, 111(1):98--136, Jan.
  2015.

\bibitem{hare2016struck}
Sam Hare, Stuart Golodetz, Amir Saffari, Vibhav Vineet, Ming-Ming Cheng,
  Stephen~L Hicks, and Philip~HS Torr.
\newblock Struck: Structured output tracking with kernels.
\newblock {\em IEEE transactions on pattern analysis and machine intelligence},
  38(10):2096--2109, 2016.

\bibitem{He2015}
Kaiming He, Xiangyu Zhang, Shaoqing Ren, and Jian Sun.
\newblock Deep residual learning for image recognition.
\newblock In {\em Proceedings of the IEEE conference on computer vision and
  pattern recognition}, pages 770--778, 2016.

\bibitem{henriques2015high}
Jo{\~a}o~F Henriques, Rui Caseiro, Pedro Martins, and Jorge Batista.
\newblock High-speed tracking with kernelized correlation filters.
\newblock {\em IEEE transactions on pattern analysis and machine intelligence},
  37(3):583--596, 2015.

\bibitem{hestenes1952methods}
Magnus~Rudolph Hestenes and Eduard Stiefel.
\newblock {\em Methods of conjugate gradients for solving linear systems},
  volume~49.
\newblock NBS Washington, DC, 1952.

\bibitem{hu2018mgcrn}
Ping Hu, Gang Wang, Xiangfei Kong, Jason Kuen, and Yap-Peng Tan.
\newblock Motion-guided cascaded refinement network for video object
  segmentation.
\newblock In {\em Proceedings of the IEEE Conference on Computer Vision and
  Pattern Recognition}, pages 1400--1409, 2018.

\bibitem{hu2018videomatch}
Yuan-Ting Hu, Jia-Bin Huang, and Alexander~G Schwing.
\newblock Videomatch: Matching based video object segmentation.
\newblock In {\em European Conference on Computer Vision}, pages 56--73.
  Springer, 2018.

\bibitem{jang2017ctn}
Won-Dong Jang and Chang-Su Kim.
\newblock Online video object segmentation via convolutional trident network.
\newblock In {\em Proceedings of the IEEE Conference on Computer Vision and
  Pattern Recognition}, pages 5849--5858, 2017.

\bibitem{johnander2018generative}
Joakim Johnander, Martin Danelljan, Emil Brissman, Fahad~Shahbaz Khan, and
  Michael Felsberg.
\newblock A generative appearance model for end-to-end video object
  segmentation.
\newblock In {\em IEEE Conference on Computer Vision and Pattern Recognition
  (CVPR)}, 2019.

\bibitem{khoreva2017lucid}
Anna Khoreva, Rodrigo Benenson, Eddy Ilg, Thomas Brox, and Bernt Schiele.
\newblock Lucid data dreaming for video object segmentation.
\newblock {\em International Journal of Computer Vision}, 127(9):1175--1197,
  Sep 2019.

\bibitem{kingma2014adam}
Diederik Kingma and Jimmy Ba.
\newblock Adam: A method for stochastic optimization.
\newblock {\em International Conference on Learning Representations}, 12 2014.

\bibitem{kjellstrom2008visual}
Hedvig Kjellstrom, Javier Romero, and Danica Kragic.
\newblock Visual recognition of grasps for human-to-robot mapping.
\newblock In {\em 2008 IEEE/RSJ International Conference on Intelligent Robots
  and Systems}, pages 3192--3199. IEEE, 2008.

\bibitem{VOT2018}
Matej Kristan, Ales Leonardis, Jiri Matas, Michael Felsberg, Roman Pfugfelder,
  Luka~Cehovin Zajc, Tomas Vojir, Goutam Bhat, Alan Lukezic, Abdelrahman
  Eldesokey, Gustavo Fernandez, and et al.
\newblock The sixth visual object tracking vot2018 challenge results.
\newblock In {\em ECCV workshop}, 2018.

\bibitem{li2018dyenet}
Xiaoxiao Li and Chen Change~Loy.
\newblock Video object segmentation with joint re-identification and
  attention-aware mask propagation.
\newblock In {\em Proceedings of the European Conference on Computer Vision
  (ECCV)}, pages 90--105, 2018.

\bibitem{lin2014microsoft}
Tsung-Yi Lin, Michael Maire, Serge Belongie, James Hays, Pietro Perona, Deva
  Ramanan, Piotr Doll{\'a}r, and C~Lawrence Zitnick.
\newblock Microsoft coco: Common objects in context.
\newblock In {\em European conference on computer vision}, pages 740--755.
  Springer, 2014.

\bibitem{luiten2018premvos}
Jonathon Luiten, Paul Voigtlaender, and Bastian Leibe.
\newblock Premvos: Proposal-generation, refinement and merging for video object
  segmentation.
\newblock In {\em Asian Conference on Computer Vision}, pages 565--580.
  Springer, 2018.

\bibitem{maninis2017osvos_s}
Kevis-Kokitsi Maninis, Sergi Caelles, Yuhua Chen, Jordi Pont-Tuset, Laura
  Leal-Taix{\'e}, Daniel Cremers, and Luc Van~Gool.
\newblock Video object segmentation without temporal information.
\newblock {\em IEEE Transactions on Pattern Analysis and Machine Intelligence
  (TPAMI)}, 2018.

\bibitem{marquardt1963algorithm}
Donald~W Marquardt.
\newblock An algorithm for least-squares estimation of nonlinear parameters.
\newblock {\em Journal of the society for Industrial and Applied Mathematics},
  11:431--441, 1963.

\bibitem{ng2002discriminative}
Andrew~Y Ng and Michael~I Jordan.
\newblock On discriminative vs. generative classifiers: A comparison of
  logistic regression and naive bayes.
\newblock In {\em Advances in neural information processing systems}, pages
  841--848, 2002.

\bibitem{oh2018rgmp}
Seoung~Wug Oh, Joon-Young Lee, Kalyan Sunkavalli, and Seon~Joo Kim.
\newblock Fast video object segmentation by reference-guided mask propagation.
\newblock In {\em 2018 IEEE/CVF Conference on Computer Vision and Pattern
  Recognition}, pages 7376--7385. IEEE, 2018.

\bibitem{oh2019video}
Seoung~Wug Oh, Joon-Young Lee, Ning Xu, and Seon~Joo Kim.
\newblock Video object segmentation using space-time memory networks.
\newblock {\em Proceedings of the IEEE International Conference on Computer
  Vision}, 2019.

\bibitem{paszke2017automatic}
Adam Paszke, Sam Gross, Soumith Chintala, Gregory Chanan, Edward Yang, Zachary
  DeVito, Zeming Lin, Alban Desmaison, Luca Antiga, and Adam Lerer.
\newblock Automatic differentiation in pytorch.
\newblock 2017.

\bibitem{perazzi2017masktrack}
Federico Perazzi, Anna Khoreva, Rodrigo Benenson, Bernt Schiele, and Alexander
  Sorkine-Hornung.
\newblock Learning video object segmentation from static images.
\newblock In {\em Proceedings of the IEEE Conference on Computer Vision and
  Pattern Recognition}, pages 2663--2672, 2017.

\bibitem{perazzi2016davis}
F. Perazzi, J. Pont-Tuset, B. McWilliams, L. {Van Gool}, M. Gross, and A.
  Sorkine-Hornung.
\newblock A benchmark dataset and evaluation methodology for video object
  segmentation.
\newblock In {\em Computer Vision and Pattern Recognition}, 2016.

\bibitem{ros2015vision}
German Ros, Sebastian Ramos, Manuel Granados, Amir Bakhtiary, David Vazquez,
  and Antonio~M Lopez.
\newblock Vision-based offline-online perception paradigm for autonomous
  driving.
\newblock In {\em 2015 IEEE Winter Conference on Applications of Computer
  Vision}, pages 231--238. IEEE, 2015.

\bibitem{ILSVRC15}
Olga Russakovsky, Jia Deng, Hao Su, Jonathan Krause, Sanjeev Satheesh, Sean Ma,
  Zhiheng Huang, Andrej Karpathy, Aditya Khosla, Michael Bernstein,
  Alexander~C. Berg, and Li Fei-Fei.
\newblock {ImageNet Large Scale Visual Recognition Challenge}.
\newblock {\em International Journal of Computer Vision (IJCV)},
  115(3):211--252, 2015.

\bibitem{saleh2016kangaroo}
Khaled Saleh, Mohammed Hossny, and Saeid Nahavandi.
\newblock Kangaroo vehicle collision detection using deep semantic segmentation
  convolutional neural network.
\newblock In {\em 2016 International Conference on Digital Image Computing:
  Techniques and Applications (DICTA)}, pages 1--7. IEEE.

\bibitem{telea2004inpainting}
Alexandru Telea.
\newblock An image inpainting technique based on the fast marching method.
\newblock {\em Journal of graphics tools}, 9(1):23--34, 2004.

\bibitem{tjaden2018region}
Henning Tjaden, Ulrich Schwanecke, Elmar Sch{\"o}mer, and Daniel Cremers.
\newblock A region-based gauss-newton approach to real-time monocular multiple
  object tracking.
\newblock {\em IEEE transactions on pattern analysis and machine intelligence},
  2018.

\bibitem{ventura2019rvos}
Carles Ventura, Miriam Bellver, Andreu Girbau, Amaia Salvador, Ferran Marques,
  and Xavier Giro-i Nieto.
\newblock Rvos: End-to-end recurrent network for video object segmentation.
\newblock In {\em Proceedings of the IEEE Conference on Computer Vision and
  Pattern Recognition}, pages 5277--5286, 2019.

\bibitem{voigtlaender2017onavos}
Paul Voigtlaender and Bastian Leibe.
\newblock Online adaptation of convolutional neural networks for video object
  segmentation.
\newblock In {\em BMVC}, 2017.

\bibitem{voigtlaender2018feelvos}
Paul Voigtlaender and Bastian Leibe.
\newblock Feelvos: Fast end-to-end embedding learning for video object
  segmentation.
\newblock In {\em IEEE Conference on Computer Vision and Pattern Recognition
  (CVPR)}, 2019.

\bibitem{vondrick2018tracking}
Carl Vondrick, Abhinav Shrivastava, Alireza Fathi, Sergio Guadarrama, and Kevin
  Murphy.
\newblock Tracking emerges by colorizing videos.
\newblock In {\em European Conference on Computer Vision}, pages 402--419.
  Springer, 2018.

\bibitem{wang2019ranet}
Ziqin Wang, Jun Xu, Li Liu, Fan Zhu, and Ling Shao.
\newblock Ranet: Ranking attention network for fast video object segmentation.
\newblock In {\em Proceedings of the IEEE International Conference on Computer
  Vision}, pages 3978--3987, 2019.

\bibitem{xu2018youtube}
Ning Xu, Linjie Yang, Yuchen Fan, Jianchao Yang, Dingcheng Yue, Yuchen Liang,
  Brian Price, Scott Cohen, and Thomas Huang.
\newblock Youtube-vos: Sequence-to-sequence video object segmentation.
\newblock In {\em European Conference on Computer Vision}. Springer, 2018.

\bibitem{xu2018youtube2}
Ning Xu, Linjie Yang, Yuchen Fan, Dingcheng Yue, Yuchen Liang, Jianchao Yang,
  and Thomas Huang.
\newblock Youtube-vos: A large-scale video object segmentation benchmark.
\newblock {\em arXiv preprint arXiv:1809.03327}, 2018.

\bibitem{yang2018osnm}
Linjie Yang, Yanran Wang, Xuehan Xiong, Jianchao Yang, and Aggelos~K
  Katsaggelos.
\newblock Efficient video object segmentation via network modulation.
\newblock In {\em Proceedings of the IEEE Conference on Computer Vision and
  Pattern Recognition}, pages 6499--6507, 2018.

\bibitem{yu2018dfn}
Changqian Yu, Jingbo Wang, Chao Peng, Changxin Gao, Gang Yu, and Nong Sang.
\newblock Learning a discriminative feature network for semantic segmentation.
\newblock In {\em Proceedings of the IEEE Conference on Computer Vision and
  Pattern Recognition}, pages 1857--1866, 2018.

\bibitem{zhao2017pyramid}
Hengshuang Zhao, Jianping Shi, Xiaojuan Qi, Xiaogang Wang, and Jiaya Jia.
\newblock Pyramid scene parsing network.
\newblock In {\em Proceedings of the IEEE conference on computer vision and
  pattern recognition}, pages 2881--2890, 2017.

\end{thebibliography}
